%% file: main.tex
\documentclass[10pt,twocolumn,letterpaper]{article}

\usepackage[pagenumbers]{cvpr} %

\input{preamble}

\definecolor{cvprblue}{rgb}{0.21,0.49,0.74}
\usepackage[pagebackref,breaklinks,colorlinks,allcolors=cvprblue]{hyperref}

\def\paperID{*****} %
\def\confName{CVPR}
\def\confYear{2027}

\title{AdaRSI: On-Demand Recursive Self-Improvement for Embodied Agents}

\author{
Peidong Wang$^{1,2}$\thanks{Work done during an internship at AgiBot.}
\quad Ying Chang$^{3}$
\quad Zhiming Ma$^{1}$
\quad Xiaocui Yang$^{1}$
\quad Shi Feng$^{1}$\textsuperscript{\dag}
\\
Daqi Liu$^{2}$
\quad Yuzhe Yang$^{2}$
\quad Xu Cao$^{2}$
\quad Chuang Wang$^{2}$
\quad Jian Su$^{2}$\textsuperscript{\dag}
\quad Daling Wang$^{1}$
\\[4pt]
$^{1}$Northeastern University
\quad $^{2}$AgiBot
\quad $^{3}$University of Chinese Academy of Sciences
}

\AddToHook{cmd/@maketitle/after}{%
  \par\vspace{-28pt}
  \begin{minipage}{\textwidth}
    \centering
    \includegraphics[width=0.95\linewidth,trim=0 15bp 0 12bp,clip]{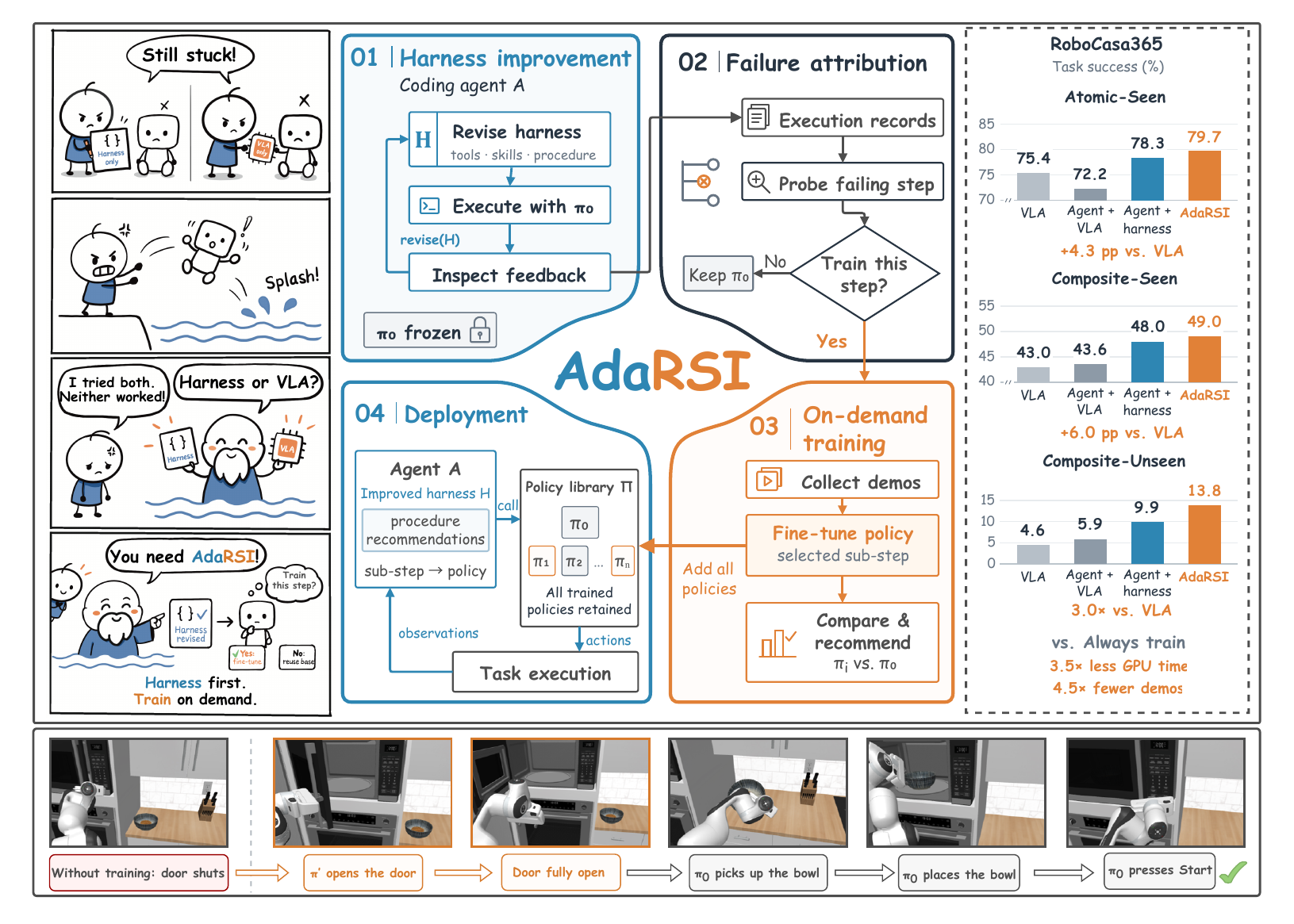}
    \captionsetup{skip=4pt}
    \captionof{figure}{\textbf{Overview of \ourmethod.} The agent improves its harness first and fine-tunes only the sub-steps that the harness cannot fix, raising success on every suite of RoboCasa365. In WaffleReheat (bottom), only the door opening is trained.}
    \label{fig:AdaRSI}
  \end{minipage}
  \par\vspace{0.5em}
}

\begin{document}
\maketitle
{\renewcommand{\thefootnote}{\dag}\footnotetext{Corresponding authors.}}

\input{sec/0_abstract}
\input{sec/1_intro}
\input{sec/2_related}
\input{sec/3_setting}
\input{sec/4_method}
\input{sec/5_experiments}
\input{sec/6_discussion}
{
    \small
    \bibliographystyle{ieeenat_fullname}
    \bibliography{main}
}

\end{document}

%% file: preamble.tex
\newcommand{\TODO}[1]{\textbf{\color{red}[TODO: #1]}}
\renewcommand{\TODO}[1]{}

\usepackage{xspace}
\newcommand{\ourmethod}{AdaRSI\xspace}

\usepackage{colortbl}
\usepackage{tabularx}
\usepackage{makecell}

\newcolumntype{C}{>{\centering\arraybackslash}X}
\usepackage{pifont}
\usepackage{pgfplots}
\pgfplotsset{compat=1.18}
\definecolor{oursblue}{HTML}{2A78D6}
\definecolor{oursrow}{HTML}{EAF2FC}
\definecolor{gaingreen}{HTML}{006300}
\definecolor{neutralgray}{HTML}{7A7A74}
\newcommand{\cmark}{{\color{oursblue}\ding{51}}}
\newcommand{\xmark}{{\color{neutralgray}\ding{55}}}

%% file: sec/0_abstract.tex
\begin{abstract}
Pretrained vision-language-action models (VLAs) often fail midway through unseen long-horizon tasks.
Recursive self-improvement (RSI) offers a remedy, but embodied RSI methods either refine the harness around a frozen executor and so cannot add a missing motion, or train the executor on persistent failures without asking whether a failure lies in how the executor is called or in what it can do.
We ask when a self-improving embodied system should change the weights of its executor, and propose \ourmethod, an on-demand RSI framework for embodied agents.
A coding agent first improves its skills and harness, then attributes persistent failures by retrying failing sub-steps from saved states, and fine-tunes only the sub-steps that remain unreliable, with demonstrations it collects itself and experience carried over from earlier tasks.
On all 50 target tasks of RoboCasa365, with GR00T~N1.5 as the executor and no human demonstrations, \ourmethod improves the VLA on all three suites and raises its success on unseen composite tasks from 4.6\% to 13.8\%.
Compared with fine-tuning every sub-step, it reaches slightly higher success while using only 28\% of the GPU time and 22\% of the self-collected demonstrations, and fine-tuning without a harness trains more yet gains far less.
\end{abstract}

%% file: sec/1_intro.tex
\section{Introduction}
\label{sec:intro}

Pretrained vision-language-action models (VLAs)~\cite{gr00t,openvla,pi0} can perform a wide range of manipulation tasks in diverse environments.
Recent models have also made progress on long-horizon tasks~\cite{mem,pi07}.
Yet on unseen multi-step tasks they still often fail midway.
A robot may approach the wrong target, misjudge whether a step has succeeded, or fail to execute a required motion.
Some of these failures lie in how the VLA, acting as the executor, is called and others in what it can do, and the failure itself does not tell which.

Recursive self-improvement (RSI), in which a system turns its own experience into persistent changes that make it better~\cite{lastai,dgm,rsisurvey}, offers a way to repair such failures without collecting human demonstrations for every new task.
An embodied system, made of a coding agent, its skills and harness, and a low-level executor such as a VLA, can improve itself at two levels.
The harness, comprising the agent's skills, its tools and the procedure that composes them, decides how the executor is called, while the executor's weights decide what it can do.

Harness-level methods refine the agent's skills and tools while keeping the executor frozen~\cite{cap,capx,aspire,shaper,rpg,harnessvla,regenharness,adahvla,robofoundry,embodiedrsi}.
Their changes are cheap, readable and easy to revert, but they stall when a task requires a motion the executor cannot perform~\cite{rpg,rsistops}.
Policy-level methods instead update the executor's weights~\cite{embodirsi,policyrsi,enpire,find,hero,appl}, some of them directing training toward tasks with low recent success~\cite{find} or toward high-risk scenarios~\cite{policyrsi}, and systems that act on both levels train on the data that a refined harness produces~\cite{harnesspai,hexaanything}.
None of these methods asks whether a persistent failure lies in how the executor is called or in a motion it lacks, although the two call for different remedies.
Training on a failure that lies in how the executor is called is not only costly but also risky, as a fine-tuned policy can perform a sub-step worse than the executor it started from.

This raises a question that current embodied RSI leaves open: when should a self-improving system change the weights of its executor?
We argue that the answer is a matter of attribution.
The weights should change only for a sub-step that stays unreliable after the agent has tried other ways of calling the executor, such as other instructions, robot poses or approaches.
Refining the harness first therefore does more than fix the failures that lie in how the executor is called.
It also sets the states from which the executor is called, so that a policy trained afterwards learns the motion that is actually missing, from the states in which it will be used.

\begin{figure}[t]
  \centering
   \includegraphics[width=0.95\linewidth]{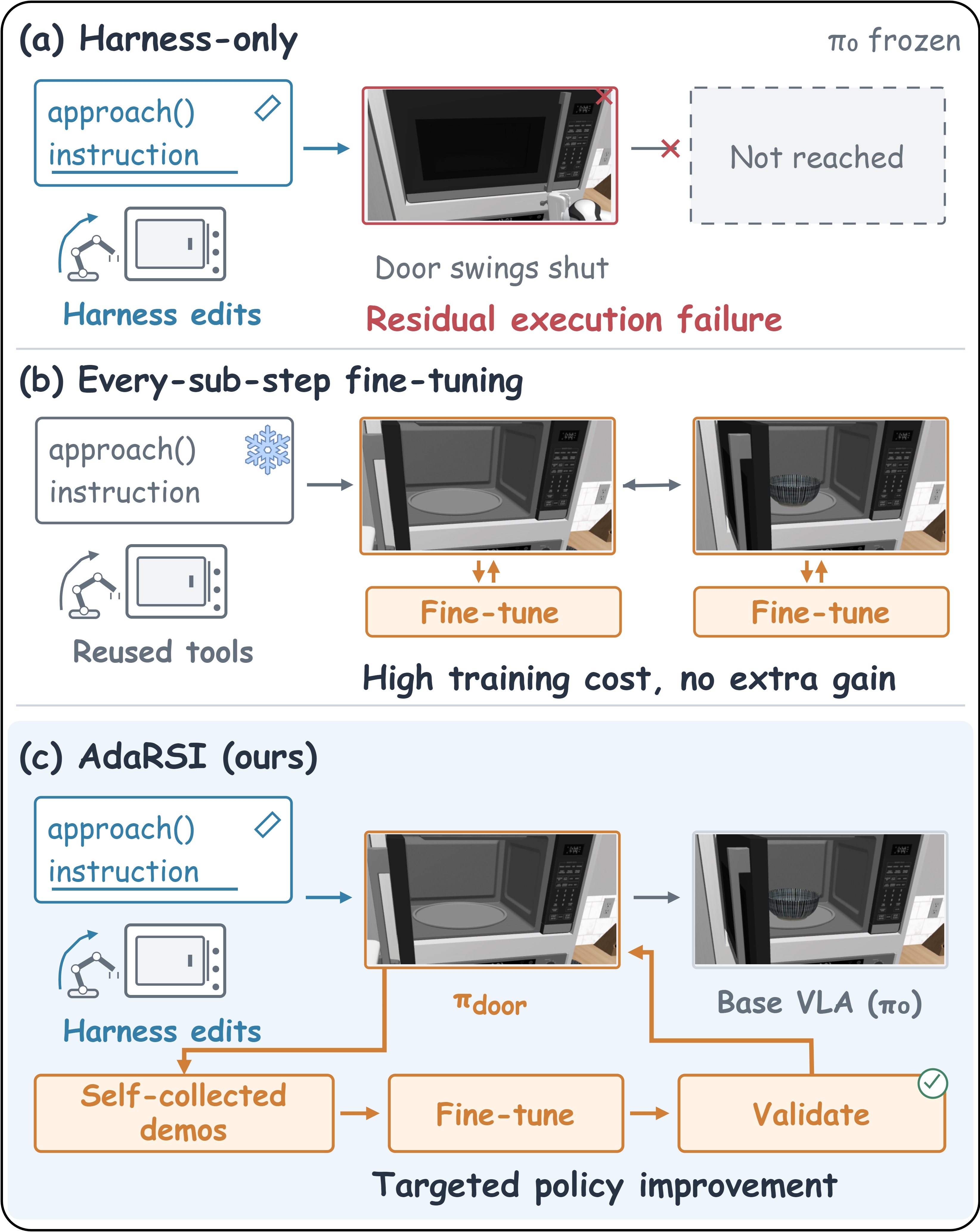}
   \caption{\textbf{Comparison of self-improvement strategies.} Harness-only improvement is bounded by what the executor can do, while fine-tuning every sub-step is costly and can make sub-steps that already worked worse. AdaRSI refines the harness first, which also sets the states from which policies are trained and called, and fine-tunes only the sub-steps that remain unreliable.}
   \label{fig:teaser}
\end{figure}

We answer this question with \textbf{\ourmethod}, an on-demand recursive self-improvement framework for embodied agents.
The agent first refines its harness from execution feedback, so that the harness absorbs the failures that better calling can fix without any training.
It then attributes the failures that persist, retrying a failing sub-step from a saved simulator state under different instructions and approaches, which separates a poor way of calling the executor from a motion it cannot make.
For the sub-steps that no such retry makes reliable, the agent collects demonstrations in simulation and fine-tunes a sub-step policy from the states its harness reaches.
Each new policy is evaluated on held-out start states, and the procedure calls it only where these measurements show that it helps, so a weight change enters the system only after it has been both attributed and validated.
Improvements accumulate over rounds, each starting from the system that the previous round produced.
Experience and a training budget derived from earlier tasks guide later ones, so the system improves not only itself but also the way it improves.
No human demonstrations are used, and privileged simulator state serves only to generate and label data, never at test time.

We evaluate on all 50 target tasks of RoboCasa365~\cite{robocasa,robocasa365} under the official protocol, with GR00T~N1.5~\cite{gr00t} as the executor.
\ourmethod improves the VLA on all three suites, most of all on unseen composite tasks, where it raises success from 4.6\% to 13.8\%, about three times the base rate.
The harness provides most of this gain, and fine-tuning, which executes only 15\% of the manipulation steps, carries the tasks that the harness alone could not solve, such as one that requires opening a microwave door the VLA never opens on its own.
In paired tests from identical states, fine-tuning teaches motions that the executor lacks, but it can degrade sub-steps that the executor already completes.
Compared with fine-tuning every sub-step and calling the new policies by default, \ourmethod reaches slightly higher success while using only 28\% of the GPU time and 22\% of the self-collected demonstrations.
Fine-tuning without a harness trains more yet gains far less, and its deployed system calls few of the policies it trained.
Training instead the sub-steps on which the executor succeeds least, under the same budget, gains nothing over the harness alone, even though most of the sub-steps it trains are the same, as attribution also tells the agent from which states a missing motion should be learned.

Our contributions are as follows:
\begin{itemize}
  \item We formulate on-demand recursive self-improvement for embodied agents, in which the decision to change the executor's weights rests on attributing each persistent failure to the harness or to the executor, and instantiate it as \ourmethod.
  \item We find that the harness decides whether training pays off, as it absorbs failures that would otherwise be trained on and sets the states from which policies are trained and called, while fine-tuning teaches motions the executor lacks but can degrade sub-steps it already completes.
  \item We show on all 50 target tasks of RoboCasa365, without any human demonstration, that training on demand improves a pretrained VLA on every suite at a fraction of the training cost of fine-tuning every sub-step.
\end{itemize}

%% file: sec/2_related.tex
\section{Related Work}
\label{sec:related}

\paragraph{Vision-language-action models.}
VLAs map images and language instructions directly to robot actions~\cite{openvla,pi0,gr00t}.
Trained on large and diverse data, they follow instructions in new environments~\cite{pi05,xr1}, carry out long-horizon tasks with memory~\cite{mem,pi07}, and can be paired with a high-level model that decomposes a task~\cite{hirobot}.
Yet they still generalize poorly to tasks outside their training distribution~\cite{robocasa365,liberopro}.
We keep a pretrained VLA as the executor and let the embodied system built on it improve itself.

\begin{figure*}[!t]
  \centering
   \includegraphics[width=0.95\linewidth]{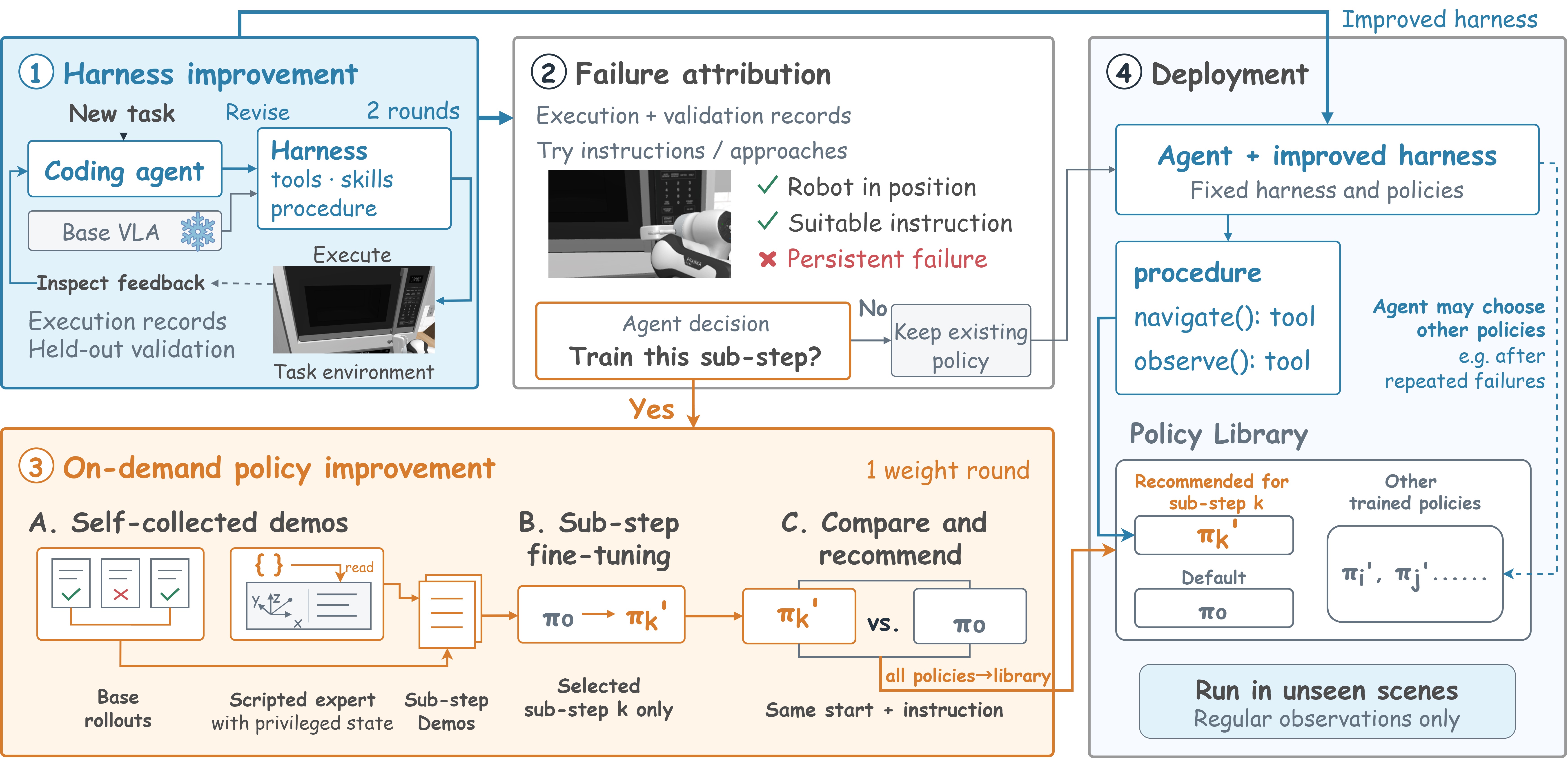}
   \caption{\textbf{Pipeline of \ourmethod.} The agent first refines its harness, then attributes persistent failures and fine-tunes only the sub-steps that remain unreliable, using demonstrations it collects itself. Trained policies join a library, from which the deployed agent calls the recommended policy for each sub-step.}
   \label{fig:pipeline}
\end{figure*}

\paragraph{Recursive self-improvement of embodied systems.}
Recursive self-improvement (RSI) refers to systems that turn their own experience into persistent changes that make them better~\cite{lastai,rsisurvey}, as language-model agents have done by rewriting their own code, harness and skills~\cite{dgm,rhi,selfharness,rsirouter}.
Most embodied methods refine the agent's skills and harness while the executor stays frozen~\cite{cap,capx,aspire,shaper,rpg,harnessvla,regenharness,adahvla,robofoundry,embodiedrsi}.
They change how and when the executor is called, but not what it can do~\cite{rsistops}.
Other methods improve the executor itself by training policies on self-collected data~\cite{embodirsi,policyrsi,enpire,find,agenticrobotics,hero,appl}.
Some of them choose where to train, practicing tasks with low recent success~\cite{find} or collecting high-risk scenarios for a specialist that a gate calls in place of the nominal policy~\cite{policyrsi}, and a few act on both levels~\cite{harnesspai,hexaanything}, for example fine-tuning the policy on data that an evolved harness collects~\cite{harnesspai}.
These methods choose which failures to train on, but not whether a failure lies in how the executor is called, which a harness change can fix, or in a motion the executor lacks, which calls for new weights.
We instead attribute each persistent failure to one of the two levels before training, and change the weights only for the sub-steps that remain unreliable after the harness has been refined.

%% file: sec/3_setting.tex
\section{Problem Formulation}
\label{sec:setting}

\paragraph{Embodied agent system.}
We consider an embodied agent system $S=(A,H,\Pi)$ with three parts.
The agent $A$ is a language-model coding agent that interprets the task, runs the system and reads the feedback from execution.
The harness $H$ is what the agent writes and uses to act, including its skills, its tools, and the program that composes them into a procedure for the task.
The executor library $\Pi$ holds the low-level policies that perform manipulation.
It starts as $\Pi_0=\{\pi_0\}$, a single pretrained VLA, and grows when a policy fine-tuned for a specific sub-step is added.
Tools may move the robot directly, for example to navigate to a fixture, but every manipulation sub-step is executed by a policy in $\Pi$, which the harness selects together with its instruction.

\paragraph{Self-improvement on a new task.}
Given a new task, the system may change itself before it is deployed.
A change acts on one of two levels, either the harness, $H \to H'$, or the executor library, $\Pi \to \Pi \cup \{\pi'\}$, where $\pi'$ is fine-tuned from a policy already in $\Pi$.
Changes are made in an improvement phase in simulation, where the system can run the task repeatedly and privileged simulator state is accessible.
No human demonstrations are available.
At deployment, the harness and the library are fixed, and the system acts from the robot's regular observations only.
Let $J(S)$ be the deployment success rate and $C(S)=C_H+C_\Pi$ the cost of improvement, where $C_H$ counts language-model tokens and $C_\Pi$ counts self-collected demonstrations and training compute, including policies that were trained but not adopted.
The goal is a system with high $J$ at low $C$.

%% file: sec/4_method.tex
\section{\ourmethod}
\label{sec:method}

\ourmethod first refines the harness and then descends to the weights.
In the harness rounds, the agent builds its tools and procedure, runs the system on held-out validation episodes, and revises the tools and procedure on this feedback.
The weight round then lets the agent descend to the executor, but only for the sub-steps that remain unreliable after these revisions.
Each round starts from the system that the previous round produced.
The harness is written by the agent for itself: at deployment, a fresh instance of the same agent reads the procedure and calls tools and policies as it prescribes.

\subsection{Harness-Level Improvement}
\label{sec:method-harness}

The harness starts minimal.
It exposes only the executor, which runs with a language instruction for a bounded number of steps, the camera images and proprioception, a map of the fixtures in the scene, low-level commands for the base and the arm, and a basic tool that navigates the base to a fixture, as a planning and control module does on a real mobile manipulator.
Everything else is written by the agent, such as tools for fine positioning in front of a fixture, resetting the arm or zooming a camera onto a fixture, among many others.
Improving the harness means writing such tools and strengthening the procedure that calls them.
The procedure tells the deployed agent which sub-steps to perform, which tool or policy to call for each, and how to verify progress.

During development, the agent may reset the scene, save and restore simulator snapshots, and query the task's success check.
Snapshots enable counterfactual re-execution, in which a single sub-step is retried from the same state under a different instruction or approach, and this separates a poor instruction or a poor approach from a motion the executor cannot make.
After each round, the system is run as at deployment on held-out validation episodes, and the records of these runs are returned to the agent as feedback for the next round.
The agent thus tunes its harness on one set of scenes and is judged on another, which exposes failures that only appear in new scenes.

\input{tables/main}

\subsection{Descending to the Weights}
\label{sec:method-weights}

\paragraph{Executor-limited sub-steps.}
Descending to the weights begins with failure attribution, which assigns each persistent failure either to the harness or to the executor.
We call a sub-step \emph{executor-limited} if no harness the agent can find lets $\pi_0$ perform it reliably.
At the end of the harness rounds, the agent records the sub-steps that still fail although the robot is in place and the instruction is one the executor knows, together with the success rate it measured for $\pi_0$ from such states.
These candidates, and the validation records behind them, guide the weight round.

\paragraph{Self-collected demonstrations.}
For a sub-step it decides to train, the agent collects demonstrations in simulation itself.
It can apply rejection sampling to the base executor, rolling it out repeatedly from the failing state and keeping the segments that succeed, or write a scripted expert that reads privileged state such as object poses and articulation states.
This follows learning with privileged information~\cite{lupi,lbc}, in which a privileged teacher labels data for a student that sees only the robot's regular observations.
Privileged state is used only here, to generate and label data, and the deployed system never sees it.

\paragraph{Fine-tuning and validation.}
Each demonstration set fine-tunes a sub-step policy, starting from $\pi_0$ or from one of the agent's earlier policies, and the result is added to $\Pi$.
The new policy is called through the same interface as the base executor.
As in self-improving coding agents~\cite{dgm}, every self-modification is validated empirically.
The agent compares the new policy with $\pi_0$ from development states and records the result in its procedure, which recommends the policy for the sub-steps where it performs better and keeps $\pi_0$ as the default everywhere else.
In the same round, the agent also works out how to call each new policy, for example how the robot is positioned before the call.
Every trained policy stays in $\Pi$, so the deployed agent follows the recommendations of its procedure but can still choose another policy from the library, for instance after $\pi_0$ fails repeatedly.

\subsection{Recursion Across Rounds and Tasks}
\label{sec:method-recursion}

Each round maps the current system and the experience gathered with it to an improved system, $S_{k+1}=g(S_k,\mathcal{E}_k)$.
The experience $\mathcal{E}_k$ contains the development runs, the validation records and, in the weight round, the training results of every policy, adopted or not.
Because the harness and the policies persist, later rounds build on what earlier rounds found, and a harness revision can make use of a newly trained policy as easily as of the base executor.

Self-improvement also carries over across tasks.
We select pilot tasks to span high, moderate, and low base-VLA success rates on manipulation sub-steps, measured on development scenes during the harness rounds.
This provides varied starting points for studying when harness refinement suffices and when weight adaptation is useful.
The pilot tasks are improved without any limit on training.
From their training logs, the agent writes an experience document that records which kinds of sub-steps benefited from training, how much data sufficed, and where effort was wasted.
The same logs set a training budget by a fixed rule, scaled from the training that the adopted policies of the pilot tasks actually used.
Every later task is improved with this document and under this budget, which the environment enforces on training steps and self-collected demonstrations.
The budget turns the choice of where to train into an explicit trade-off, and the experience tells the agent which sub-steps are likely to be executor-limited.
In this sense the system improves not only itself but also the way it improves.

At deployment, a fresh instance of the agent runs the final system on unseen scenes.
It follows the procedure, calls tools and policies, and checks progress from images, without snapshots, success checks or privileged state.

%% file: tables/main.tex
\begin{table*}[t]
\centering
\begingroup
\small
\definecolor{tableaccent}{HTML}{245E86}
\definecolor{tablelight}{HTML}{F2F6FA}
\definecolor{tableours}{HTML}{E4EEF8}
\definecolor{tablemuted}{HTML}{797979}
\definecolor{tablegain}{HTML}{3B7198}
\newcommand{\tabaccent}[1]{\textcolor{tableaccent}{#1}}
\newcommand{\tabgain}[1]{\textcolor{tablegain}{#1}}
\newcommand{\tabmuted}[1]{\textcolor{tablemuted}{#1}}
\renewcommand{\cmark}{{\color{tableaccent}\ding{51}}}
\renewcommand{\xmark}{{\color{tablemuted}\ding{55}}}
\setlength{\tabcolsep}{3pt}
\renewcommand{\arraystretch}{1.13}
\begin{tabularx}{\textwidth}{l ccc CCCC}
\toprule
 & \multicolumn{3}{c}{Self-improvement} & \multicolumn{4}{c}{Success rate on RoboCasa365 (\%)} \\
\cmidrule(lr){2-4} \cmidrule(l){5-8}
\textbf{Method} & Agent & Harness & Weights
 & \makecell{Atomic-Seen\\[-1pt]{\footnotesize\color{tablemuted}18 tasks}}
 & \makecell{Composite-Seen\\[-1pt]{\footnotesize\color{tablemuted}16 tasks}}
 & \makecell{Composite-Unseen\\[-1pt]{\footnotesize\color{tablemuted}16 tasks}}
 & \makecell{All\\[-1pt]{\footnotesize\color{tablemuted}50 tasks}} \\
\midrule
GR00T N1.5~\cite{gr00t} & \xmark & \xmark & \xmark & 75.4 & 43.0 & 4.6 & 42.4 \\
Agent + VLA             & \cmark & \xmark & \xmark & 72.2 & 43.6 & 5.9 & 41.8 \\
\rowcolor{tablelight}
Agent + harness         & \cmark & \cmark & \xmark & 78.3 & 48.0 & 9.9 & 46.7 \\
\midrule
\rowcolor{tableours}
\tabaccent{\textbf{\ourmethod (ours)}} & \cmark & \cmark & \cmark & \textbf{79.7} & \textbf{49.0} & \textbf{13.8} & \tabaccent{\textbf{48.8}} \\
{\footnotesize\color{tablemuted}\quad Gain over VLA} & & & & {\footnotesize\color{tablegain}+4.3} & {\footnotesize\color{tablegain}+6.0} & {\footnotesize\color{tablegain}+9.2} & {\footnotesize\color{tablegain}+6.4} \\
\bottomrule
\end{tabularx}
\endgroup
\caption{\textbf{Success rates (\%) on the 50 target tasks of RoboCasa365.} Each task is tested on 50 scenes, \emph{All} averages over the 50 tasks, and no system uses human demonstrations.}
\label{tab:main}
\end{table*}

%% file: sec/5_experiments.tex
\section{Experiments}
\label{sec:exp}

\subsection{Setup}
\label{sec:exp-setup}

\paragraph{Benchmark.}
We evaluate on all 50 target tasks of RoboCasa365~\cite{robocasa365}, which form three suites.
The 18 Atomic-Seen and 16 Composite-Seen tasks appear in the executor's training data, while the 16 Composite-Unseen tasks compose familiar sub-tasks into tasks that are absent from it.
We follow the official protocol, with the target split, whose kitchens and objects do not overlap with pretraining, the official task horizons, which also count the steps taken by tools, and the official success check.
Each task is tested on 50 scenes.
Development, validation and test scenes are disjoint, and test scenes are never used during improvement.
We report the success rate of each task, averaged over the tasks of a suite.

\paragraph{Implementation.}
The executor $\pi_0$ is GR00T~N1.5~\cite{gr00t}, using the official Composite-Seen checkpoint for composite tasks and the Atomic-Seen checkpoint for atomic tasks.
The agent is GPT-6.1-sol, run through Codex.
Pilot tasks in each suite provide the experience document and the training budget described in \cref{sec:method-recursion}.

\input{tables/unseen}

\paragraph{Compared systems.}
We compare four systems.
(1) \emph{VLA} runs $\pi_0$ alone with the full task instruction, as in the official protocol.
(2) \emph{Agent + VLA} lets the agent call $\pi_0$ with its own instructions, without development or training.
(3) \emph{Agent + harness} stops after the harness rounds.
(4) \emph{\ourmethod} adds the weight round to this system.
Further variants that remove or replace parts of \ourmethod are introduced with the ablations in \cref{sec:exp-ablation}.

\subsection{Main Results}
\label{sec:exp-main}

\paragraph{Overall.}
\ourmethod achieves the highest suite-level success rate on all three suites, reaching 48.8\% over the 50 tasks compared with 42.4\% for the VLA alone and 46.7\% for \emph{Agent + harness} (\cref{tab:main}).
The largest gain is on Composite-Unseen, where success rises from 4.6\% to 13.8\%, about three times the base executor's rate.
The gains also extend to tasks present in the executor's training data, with Atomic-Seen rising from 75.4\% to 79.7\% and Composite-Seen from 43.0\% to 49.0\%.
The benefit is thus not confined to acquiring motions for unseen tasks, and improving how existing capabilities are used also raises task success.

\paragraph{Harness level.}
Letting the agent call the VLA without any development does not reliably raise success.
\emph{Agent + VLA} falls from 75.4\% to 72.2\% on Atomic-Seen, and its overall success remains below that of the VLA alone.
Harness development reverses this drop and improves all three suites, reaching 78.3\% on Atomic-Seen and more than doubling success on Composite-Unseen to 9.9\%.
In CuttingToolSelection, for example, the agent writes a rule that maps each food to the right tool instead of leaving ``the appropriate cutting tool'' to the VLA, and adds checks that stop a failing attempt early, raising success from 30\% to 50\%.
The benefit therefore comes from improving the tools and procedure through execution feedback, rather than simply allowing an agent to issue sub-instructions.

\paragraph{Weight level.}
The weight round adds the most on Composite-Unseen, a further 3.9 percentage points against 1.4 on Atomic-Seen and 1.0 on Composite-Seen.
\Cref{tab:unseen} breaks this gain down by task.
WaffleReheat shows the largest effect, rising from 0\% with the harness alone to 38\% with \ourmethod.
Its harness reaches the microwave, but $\pi_0$ cannot open the door and release the handle, and a policy fine-tuned for this sub-step resolves the bottleneck while $\pi_0$ still performs every other step.
RecycleBottlesByType improves from 0\% to 18\% and BreadSelection from 42\% to 52\%, so the weight round helps both tasks that rarely succeed and tasks where the harness already does well.
Not every task benefits, however.
CuttingToolSelection falls from 50\% to 36\% after the weight round, and several tasks remain unsolved by every system although the agent fine-tunes policies for their sub-steps.
We return to these cases in \cref{sec:discussion}.

\paragraph{Where the weights change.}
\input{tables/adoption}

\Cref{tab:adoption} shows where \ourmethod changes the weights.
On the two seen suites, whose tasks $\pi_0$ was trained on, most trained policies are never called.
The deployed systems use only 12 of 72 policies on Atomic-Seen and 8 of 34 on Composite-Seen, and $\pi_0$ still executes 78\% and 98\% of their manipulation steps.
On Atomic-Seen this share counts whole episodes, since an atomic task has a single sub-step.

On Composite-Unseen, where $\pi_0$ lacks the motions that the new tasks require, 32 of 38 policies are used across all 16 tasks, and 85\% of the training time goes into them.
Training thus concentrates where the executor falls short.
Even so, each policy replaces only the sub-step it was trained for, and fine-tuned policies execute 15\% of the manipulation steps, against 71\% for \emph{Always train}, which spends 3.5$\times$ the GPU time.
In WaffleReheat, for instance, the fine-tuned policy opens the microwave door and $\pi_0$ performs every other sub-step.

A paired test from identical states isolates what the fine-tuned weights themselves change (\cref{tab:substep}).
We replay a test episode up to the first call of a policy, save a snapshot, and let $\pi_0$ and the fine-tuned policy execute the same instruction for the same number of steps, with privileged state judging success.
On motions beyond $\pi_0$, fine-tuning teaches what is missing.
$\pi_0$ opens the microwave door in none of 18 kitchens and the fine-tuned policy in 7, and the mean opening rises from 0.01 to 0.61, while the glass bottle is picked up in 17 of 44 starts against 2.
On motions that $\pi_0$ already performs, such as placing the croissant or the jam on the cutting board, fine-tuning lowers success.
The same training thus helps where the executor falls short and hurts where it does not, which is the case for training on demand.

\input{tables/substep}

\subsection{Ablations}
\label{sec:exp-ablation}

We ablate \ourmethod on the Composite-Unseen suite with five variants.
(1) \emph{Agent + weights} improves the system only by fine-tuning, without harness rounds, tools or a training budget.
(2) \emph{Agent + harness} improves only the harness and stops before the weight round.
(3) \emph{Always train} fine-tunes a policy for every manipulation sub-step and calls the fine-tuned policies by default.
(4) \emph{Fixed procedure} runs the procedure and policies that \ourmethod builds as a fixed program, without the agent at deployment.
(5) \emph{\ourmethod w/ $\pi_0$} keeps the final system of \ourmethod but runs $\pi_0$ wherever a fine-tuned policy would be called.

\input{tables/ablation}

\paragraph{Success against cost.}
\Cref{fig:pareto} places every system by its training cost and its success.
The arrows trace the order in which \ourmethod improves the system.
Harness rounds raise success from 5.9\% to 9.9\% without any training, and the weight round then adds 3.9 points for 966 GPU minutes.
Every other system that trains lands in the region that \ourmethod dominates, with at least the same training and lower success.
\emph{Agent + weights} trains more and reaches 7.2\%, \emph{Always train} spends 3.5$\times$ the GPU time for 12.6\%, and \emph{Fixed procedure} uses exactly the same policies for 6.9\%.
The systems that train nothing are cheaper, but none of them exceeds 9.9\%, so no system we compare is both cheaper and more successful than \ourmethod.

\input{figures/pareto}

\paragraph{Neither level suffices alone.}
Both levels help on their own, but neither reaches their combination.
Fine-tuning without a harness lifts the VLA from 4.6\% to 7.2\%, and the harness alone reaches 9.9\%, while \ourmethod reaches 13.8\% on the same episodes.
The harness is the larger lever, since most failures come from the procedure rather than from missing motions.
It also does more than raise success, because it tells the agent where training is needed.
Without harness rounds, the agent trains more than \ourmethod, 45 policies against 38 and 1.25$\times$ the GPU time, yet the deployed system calls only 5 of them, against 32 for \ourmethod (\cref{tab:adoption}).
Many of these policies take on work that a tool would do, such as moving the base or carrying an object across the kitchen, and others are trained from wherever the robot happens to stand.
In WaffleReheat, for instance, both systems train a policy for the microwave door, but only \ourmethod trains it from the pose to which its tools bring the robot, so only its policy is ever used, and the task succeeds in 38\% of episodes against 4\% for the agent without harness rounds.
The harness thus attributes failures to the executor and fixes the states from which a policy is trained and called, which is what makes training on demand effective.
\input{tables/swap}

\input{tables/every}

\paragraph{Weights and procedure.}
The weight round changes both the policies and the procedure that calls them.
\emph{\ourmethod w/ $\pi_0$} separates the two on the five tasks whose success changes in that round, so that \cref{tab:swap}, read from left to right, adds first the revised procedure and then the fine-tuned weights.
Of the 31 successes that \ourmethod adds to the harness on these tasks, 17 disappear when $\pi_0$ replaces the fine-tuned policies, and the other 14 come from the revised procedure.
The weights matter most where $\pi_0$ lacks the motion.
WaffleReheat falls from 38\% to 10\% once $\pi_0$ has to open the microwave door again, while BreadSelection keeps its whole gain, which comes from revised instructions.
The loss in CuttingToolSelection already appears with the revised procedure alone, which we return to in \cref{sec:discussion}.
Fine-tuning and the procedure built around it thus contribute about equally, and the weights carry the tasks that no procedure can solve.

\input{tables/selection}

\paragraph{Training every sub-step.}

Fine-tuning every manipulation sub-step does not improve on \ourmethod (\cref{tab:every}).
It reaches 12.6\% against 13.8\% on the same episodes, while it trains 3.1$\times$ as many policies, collects 4.5$\times$ as many demonstrations and uses 3.5$\times$ the GPU time.
The task-level results show why.
In WaffleReheat, training everywhere replaces every sub-step, including picking up the bowl, which $\pi_0$ already performs reliably, and the new policy does it worse, so success drops from 38\% to 28\%.
In WashFruitColander, most of the twelve new policies do not beat $\pi_0$ on held-out starts, and success drops from 58\% to 34\%, partly because calling many policies also slows each episode until some reach the time limit.
Training everywhere helps where \ourmethod left a useful sub-step untrained, most clearly in RecycleBottlesByType, where policies that carry and place each bottle raise success from 18\% to 30\%.
Overall, most of the extra training buys nothing, and training everywhere is not an appropriate default, because a fine-tuned policy can replace an executor call that already worked.
\paragraph{Choosing what to train.}
A simpler rule could decide where to train without attributing failures, by training the sub-steps on which $\pi_0$ succeeds least.
\emph{Lowest first} applies this rule on the Composite-Unseen tasks that are not pilots.
It starts from the same harness as \ourmethod, with the same budget, demonstration collection and adoption rule, measures $\pi_0$ on every manipulation sub-step from the states that the procedure reaches, and trains as many sub-steps per task as \ourmethod does, lowest success first, without retrying any of them from other states or with other instructions.
It gains nothing over the harness alone, reaching 11.8\% against 12.0\%, while \ourmethod reaches 16.6\% at a similar training cost (\cref{tab:selection}).
Where it trains, the rule mostly selects the same sub-steps as \ourmethod, 11 of the 13 it trains, yet it does not turn them into gains.
In WaffleReheat, both systems train a policy for the microwave door, but only \ourmethod, which retries the failing sub-step before training, finds that its failure depends on how the hand approaches the handle and trains from an approach that works, so the task succeeds in 38\% of episodes against 2\%.
Measuring every sub-step also takes time, and in four tasks it used up the development session before any policy was trained.
Attribution thus tells the agent not only which sub-steps to train but also from which states to learn them.

\input{tables/outcomes}

\paragraph{Adaptive execution at deployment.}
At deployment, the agent uses execution feedback to adjust how it calls tools and policies, revising sub-goals, instructions and recovery actions as the task unfolds.
The fixed-procedure variant also responds to feedback, but its choices are limited to the conditions and responses encoded in advance.
To evaluate the value of this adaptation, we execute the final procedure and policies of \ourmethod as a fixed program, with privileged simulator state available for state estimation.
The fixed program reaches 6.9\% success against 13.8\% for \ourmethod, despite this information advantage (\cref{tab:ablation}).
The largest drops occur on tasks that require choosing a standing position from the current view or changing approach after a failure, as WaffleReheat falls from 38\% to 4\% and WashFruitColander from 58\% to 28\% (\cref{tab:unseen}).
By contrast, on CuttingToolSelection, whose procedure follows fixed steps, the program reaches 40\% against 36\% for \ourmethod.
These results support keeping the agent in the execution loop.
It adjusts how it uses its tools and policies to the observed state, beyond the responses predefined in the fixed program.

\subsection{Cost}
\label{sec:exp-cost}

On the Composite-Unseen suite, \ourmethod trains 2.4 policies per task on average, adopted or not, using 26 self-collected demonstrations, 1.9k updates and about one GPU-hour.
Its improvement phase, including the harness rounds, validation and the weight round, consumes 154M input tokens and 0.64M output tokens per task, and at deployment the agent uses 4.2M input and 7.2k output tokens per episode, with about 97\% of the input tokens being cache hits.
The improvement phase costs 117M and 137M input tokens per task on Atomic-Seen and Composite-Seen, where deployment needs 1.2M and 1.9M per episode, since the procedures are shorter.
Fine-tuning every sub-step raises the deployment cost to 5.3M input tokens per episode, because the agent calls and checks more policies, while fine-tuning without a harness needs fewer tokens per episode, 1.5M, but fails more often.
No human demonstration is used at any stage.

\input{figures/case_study_fig}

\subsection{Case Studies}
\label{sec:exp-cases}
\label{sec:exp-inside}

\paragraph{During training.}
To see how the agent decides where to train, we analyze the weight rounds of the pilot tasks, which run without a training budget and whose records produce the experience document.
\Cref{tab:outcomes} sorts representative policies by whether $\pi_0$ could already perform the sub-step and whether fine-tuning improved it on held-out starts.
When a sub-step is beyond $\pi_0$, a few demonstrations can teach it, as twenty demonstrations do for the left door of a French-door refrigerator in GarnishPancake.
The task still fails there, because later sub-steps remain unsolved, so learning one motion does not by itself make a long task succeed.
Other missing motions are learned only partly, such as picking up the skewer, or not at all, such as closing the toaster-oven door.
When $\pi_0$ already performs a sub-step, training can still raise its reliability, as for the mug in CoffeeSetupMug, but it often leaves the sub-step unchanged or yields a gain in development that does not hold at test, as for the drawer in OpenDrawer.
The experience document summarizes such cases for the tasks that follow.

\paragraph{During deployment.}
\Cref{fig:case} follows two tasks through the system and shows when each level of improvement is the right one.
In WashFruitColander (\cref{fig:case}a), the VLA puts fruit into the sink before placing the colander, then leaves the fruit outside it.
The harness instead places the colander first and puts in each fruit with a short instruction.
When the faucet attempts produce no water, the agent inspects the scene and uses its working-pose tool to reposition the base before retrying, completing the task without training any policy.
GetToastedBread (\cref{fig:case}b) fails for a different reason.
After the toast pops up, the base VLA rolls the wrist without grasping the bread, and resetting the arm and re-parking the base before retrying still leave it in the toaster.
\ourmethod uses a policy fine-tuned for bread pickup with the same instruction, retrieving the bread in two 160-step calls.
The robot then navigates with the bread held, and $\pi_0$ places it on the plate.
The two tasks thus call for different levels of improvement, and \ourmethod applies each only where it is needed.

%% file: tables/unseen.tex
\begin{table*}[t]
\centering
\begingroup
\small
\definecolor{unseenaccent}{HTML}{245E86}
\definecolor{unseenlight}{HTML}{F2F6FA}
\definecolor{unseenfixed}{HTML}{FFFFFF}
\definecolor{unseenwash}{HTML}{E4EEF8}
\definecolor{unseenpositive}{HTML}{3B7198}
\definecolor{unseenmuted}{HTML}{797979}
\newcommand{\unseengain}[1]{\textcolor{unseenpositive}{#1}}
\newcommand{\unseendecline}[1]{\textcolor{unseenmuted}{#1}}
\newcommand{\unseenzero}[1]{\textcolor{unseenmuted}{#1}}
\newcommand{\unseenmeanvalue}[1]{\textcolor{unseenaccent}{#1}}
\setlength{\tabcolsep}{3pt}
\renewcommand{\arraystretch}{1.13}
\begin{tabularx}{\textwidth}{@{}l C C >{\columncolor{unseenfixed}}C >{\columncolor{unseenlight}}C >{\columncolor{unseenlight}}C >{\columncolor{unseenlight}}C >{\columncolor{unseenwash}}C C@{}}
\toprule
 & \multicolumn{2}{c}{No self-improvement} & \multicolumn{1}{>{\columncolor{unseenfixed}}c}{\footnotesize No agent} & \multicolumn{4}{>{\columncolor{unseenlight}}c}{\textcolor{unseenaccent}{Self-improving agents}} & \\
\cmidrule(lr){2-3} \cmidrule(lr){4-4} \cmidrule(lr){5-8}
\textbf{Task} & VLA & \makecell{Agent +\\VLA} & \makecell{Fixed\\procedure} & \textcolor{unseenaccent}{\makecell{Agent +\\weights}} & \textcolor{unseenaccent}{\makecell{Agent +\\harness}} & \textcolor{unseenaccent}{\makecell{Always\\train}} & \textcolor{unseenaccent}{\textbf{\ourmethod}} & \makecell{$\Delta$ vs.\\harness} \\
\midrule
ArrangeBreadBasket & 0.0 & 0.0 & 0.0 & \textbf{2.0} & \textbf{2.0} & 0.0 & \textbf{2.0} & \unseenzero{0.0} \\
ArrangeTea & 0.0 & 0.0 & 0.0 & 0.0 & \textbf{2.0} & \textbf{2.0} & 0.0 & \unseendecline{-2.0} \\
BreadSelection & 22.0 & 22.0 & 36.0 & 26.0 & 42.0 & \textbf{56.0} & 52.0 & \unseengain{+10.0} \\
CategorizeCondiments & 0.0 & 0.0 & 0.0 & 0.0 & 0.0 & \textbf{6.0} & 2.0 & \unseengain{+2.0} \\
CuttingToolSelection & 20.0 & 30.0 & 40.0 & 42.0 & \textbf{50.0} & 32.0 & 36.0 & \unseendecline{-14.0} \\
GarnishPancake & 0.0 & 0.0 & 0.0 & 0.0 & 0.0 & \textbf{2.0} & 0.0 & \unseenzero{0.0} \\
GatherTableware & 2.0 & 2.0 & 2.0 & 2.0 & 6.0 & 6.0 & \textbf{8.0} & \unseengain{+2.0} \\
HeatKebabSandwich & 0.0 & 0.0 & 0.0 & 0.0 & 0.0 & 0.0 & 0.0 & \unseenzero{0.0} \\
MakeIceLemonade & 0.0 & 0.0 & 0.0 & 0.0 & 0.0 & 0.0 & 0.0 & \unseenzero{0.0} \\
PanTransfer & 0.0 & 0.0 & 0.0 & 0.0 & 2.0 & 2.0 & \textbf{4.0} & \unseengain{+2.0} \\
PortionHotDogs & 2.0 & 2.0 & 0.0 & \textbf{6.0} & 4.0 & 4.0 & 0.0 & \unseendecline{-4.0} \\
RecycleBottlesByType & 0.0 & 0.0 & 0.0 & 0.0 & 0.0 & \textbf{30.0} & 18.0 & \unseengain{+18.0} \\
SeparateFreezerRack & 0.0 & 0.0 & 0.0 & 0.0 & 0.0 & 0.0 & 0.0 & \unseenzero{0.0} \\
WaffleReheat & 2.0 & 0.0 & 4.0 & 4.0 & 0.0 & 28.0 & \textbf{38.0} & \unseengain{+38.0} \\
WashFruitColander & 26.0 & 38.0 & 28.0 & 34.0 & 48.0 & 34.0 & \textbf{58.0} & \unseengain{+10.0} \\
WeighIngredients & 0.0 & 0.0 & 0.0 & 0.0 & \textbf{2.0} & 0.0 & \textbf{2.0} & \unseenzero{0.0} \\
\midrule
\textbf{Mean} & 4.6 & 5.9 & 6.9 & 7.2 & 9.9 & 12.6 & \unseenmeanvalue{\textbf{13.8}} & \unseenmeanvalue{\textbf{+3.9}} \\
\bottomrule
\end{tabularx}
\endgroup
\caption{\textbf{Per-task success rates (\%) on Composite-Unseen.} All systems, including the ablations of \cref{sec:exp-ablation}, run on the same 50 scenes per task. The last column is the gain of \ourmethod over \emph{Agent + harness} in points.}
\label{tab:unseen}
\end{table*}

%% file: tables/adoption.tex
\begin{table}[t]
\centering
\begingroup
\small
\definecolor{tableaccent}{HTML}{245E86}
\definecolor{tableours}{HTML}{E4EEF8}
\definecolor{tablewarning}{HTML}{A65345}
\definecolor{tablerule}{HTML}{B5BBC1}
\newcommand{\tabaccent}[1]{\textcolor{tableaccent}{#1}}
\newcommand{\tabwarning}[1]{\textcolor{tablewarning}{#1}}
\setlength{\tabcolsep}{3pt}
\renewcommand{\arraystretch}{1.15}
\begin{tabularx}{\columnwidth}{l CCCC}
\toprule
 & \multicolumn{1}{c}{\footnotesize Policies} & \multicolumn{2}{c}{\footnotesize Training} & \multicolumn{1}{c}{\footnotesize Execution} \\
\cmidrule(lr){2-2} \cmidrule(lr){3-4} \cmidrule(lr){5-5}
\textbf{Suite / variant} & \makecell{Used /\\trained} & \makecell{GPU\\min.} & \makecell{Used\\share} & \makecell{FT\\steps} \\
\midrule
Atomic-Seen & 12/72 & 1,175 & 17\% & 21.7\% \\
Composite-Seen & 8/34 & 901 & 25\% & 2.3\% \\
\addlinespace[3pt]
\multicolumn{5}{l}{\hbox to \dimexpr\columnwidth-2\tabcolsep\relax{\footnotesize\textbf{Composite-Unseen}\hspace{5pt}{\color{tablerule}\leaders\hrule height3.3pt depth-3pt\hfill}}} \\
\addlinespace[2pt]
Agent + weights & \tabwarning{\textbf{5/45}} & 1,204 & \tabwarning{\textbf{12\%}} & 3.9\% \\
Always train & 104/118 & \tabwarning{\textbf{3,411}} & 89\% & \tabwarning{\textbf{71.2\%}} \\
\addlinespace[2pt]
\rowcolor{tableours}
\tabaccent{\textbf{\ourmethod}} & \tabaccent{\textbf{32/38}} & \tabaccent{\textbf{966}} & \tabaccent{\textbf{85\%}} & \tabaccent{\textbf{14.7\%}} \\
\bottomrule
\end{tabularx}
\endgroup
\caption{\textbf{Policies trained and used at test time.} A policy counts as used if the deployed agent calls it, used share is the share of training time spent on used policies, and FT steps is the share of manipulation steps at test time that fine-tuned policies execute. The counts of Always train include the policies of \ourmethod, which it reuses.}
\label{tab:adoption}
\end{table}

%% file: tables/substep.tex
\begin{table}[t]
\centering
\begingroup
\small
\definecolor{tableaccent}{HTML}{245E86}
\definecolor{tablewarning}{HTML}{A65345}
\definecolor{tablemuted}{HTML}{797979}
\definecolor{tablegroup}{HTML}{F5F5F5}
\newcommand{\tabaccent}[1]{\textcolor{tableaccent}{#1}}
\newcommand{\tabwarning}[1]{\textcolor{tablewarning}{#1}}
\renewcommand{\arraystretch}{1.13}
\setlength{\tabcolsep}{4pt}
\begin{tabularx}{\columnwidth}{@{}l C >{\centering\arraybackslash}m{45pt} >{\centering\arraybackslash}m{20pt}@{}}
\toprule
\textbf{Sub-step} & $\pi_0$ & Fine-tuned & $\Delta$ \\
\midrule
\multicolumn{4}{@{}>{\columncolor{tablegroup}[0pt][0pt]}l@{}}{\rule{0pt}{11pt}\footnotesize\textbf{Motions beyond $\pi_0$}\rule[-3pt]{0pt}{0pt}} \\
\addlinespace[2pt]
\makecell[l]{Open microwave door\\[-1pt]{\footnotesize\color{tablemuted}WaffleReheat}} & 0/18 & \tabaccent{\textbf{7/18}} & \tabaccent{\textbf{+7}} \\
\makecell[l]{Pick up glass bottle\\[-1pt]{\footnotesize\color{tablemuted}RecycleBottlesByType}} & 2/44 & \tabaccent{\textbf{17/44}} & \tabaccent{\textbf{+15}} \\
\addlinespace[5pt]
\multicolumn{4}{@{}>{\columncolor{tablegroup}[0pt][0pt]}l@{}}{\rule{0pt}{11pt}\footnotesize\textbf{Motions $\pi_0$ already performs}\rule[-3pt]{0pt}{0pt}} \\
\addlinespace[2pt]
\makecell[l]{Place croissant on board\\[-1pt]{\footnotesize\color{tablemuted}BreadSelection}} & \textbf{16/40} & \tabwarning{11/40} & \tabwarning{\textbf{-5}} \\
\makecell[l]{Place jam on board\\[-1pt]{\footnotesize\color{tablemuted}BreadSelection}} & \textbf{14/39} & \tabwarning{9/39} & \tabwarning{\textbf{-5}} \\
\bottomrule
\end{tabularx}
\endgroup
\caption{\textbf{Paired sub-step test.} $\pi_0$ and the fine-tuned policy run from the same states with the same instruction and number of steps, and entries give successful over tested starts.}
\label{tab:substep}
\end{table}

%% file: tables/ablation.tex
\begin{table*}[t]
\centering
\begingroup
\footnotesize
\definecolor{tableaccent}{HTML}{245E86}
\definecolor{tablelight}{HTML}{F2F6FA}
\definecolor{tableours}{HTML}{E4EEF8}
\definecolor{tablemuted}{HTML}{797979}
\newcommand{\tabaccent}[1]{\textcolor{tableaccent}{#1}}
\newcommand{\tabmuted}[1]{\textcolor{tablemuted}{#1}}
\renewcommand{\cmark}{{\color{tableaccent}\ding{51}}}
\renewcommand{\xmark}{{\color{tablemuted}\ding{55}}}
\setlength{\tabcolsep}{2.5pt}
\renewcommand{\arraystretch}{1.13}
\begin{tabularx}{\textwidth}{l >{\centering\arraybackslash}m{22pt} >{\centering\arraybackslash}m{29pt} >{\centering\arraybackslash}m{28pt} >{\centering\arraybackslash}m{39pt} C CCC}
\toprule
 & \multicolumn{4}{c}{Self-improvement} & & \multicolumn{3}{c}{Training cost (16 tasks)} \\
\cmidrule(lr){2-5} \cmidrule(l){7-9}
\textbf{Method} & \makecell{Online\\agent} & Harness & Weights & On-demand & \tabaccent{\makecell{Success\\(\%)}} & Policies & Demos & \makecell{GPU\\min.} \\
\midrule
GR00T N1.5~\cite{gr00t} & \xmark & \xmark & \xmark & -- & 4.6 & 0 & 0 & 0 \\
Fixed procedure         & \xmark & \cmark & \cmark & \cmark & 6.9 & \multicolumn{3}{c}{\tabmuted{reused from \ourmethod}} \\
\addlinespace[2pt]
\rowcolor{tablelight}
Agent + VLA                & \cmark & \xmark & \xmark & -- & 5.9 & 0 & 0 & 0 \\
\rowcolor{tablelight}
Agent + weights            & \cmark & \xmark & \cmark & \cmark & 7.2 & 45 & 556 & 1,204 \\
\rowcolor{tablelight}
Agent + harness            & \cmark & \cmark & \xmark & -- & 9.9 & 0 & 0 & 0 \\
\rowcolor{tablelight}
Always train          & \cmark & \cmark & \cmark & \xmark & 12.6 & 118 & 1,877 & 3,411 \\
\addlinespace[2pt]
\rowcolor{tableours}
\tabaccent{\textbf{\ourmethod}} & \cmark & \cmark & \cmark & \cmark & \tabaccent{\textbf{13.8}} & \tabaccent{38} & \tabaccent{416} & \tabaccent{966} \\
\bottomrule
\end{tabularx}
\endgroup
\caption{\textbf{Ablations on the 16 Composite-Unseen tasks.} \emph{Online agent} means that the agent runs at deployment, and \emph{on-demand} that it fine-tunes only selected sub-steps. Training cost counts every policy trained, and that of \emph{Always train} includes the training of \ourmethod.}
\label{tab:ablation}
\end{table*}

%% file: figures/pareto.tex
\begin{figure}[t]
\centering
\begingroup
\definecolor{psaccent}{HTML}{245E86}
\definecolor{psvariant}{HTML}{7D8D97}
\definecolor{pswarm}{HTML}{7D8D97}
\definecolor{psmuted}{HTML}{70777E}
\definecolor{psfixed}{HTML}{A65345}
\definecolor{psgrid}{HTML}{E2E7EB}
\definecolor{psbase}{HTML}{69869C}
\definecolor{psstage}{HTML}{4D7FA3}
\definecolor{pspath}{HTML}{4D7FA3}
\begin{tikzpicture}
\begin{axis}[
  width=\columnwidth, height=0.68\columnwidth,
  xmin=-200, xmax=3800, ymin=4, ymax=16,
  xlabel={Training GPU time (min)}, ylabel={Success (\%)},
  label style={font=\small,text=black!85},
  tick label style={font=\footnotesize,text=black!65},
  axis lines*=left, axis line style={psmuted!75,line width=0.45pt},
  tick style={psmuted!70,line width=0.4pt}, tick align=outside,
  ymajorgrids, grid style={psgrid,line width=0.35pt},
  xtick={0,1000,2000,3000}, ytick={4,8,12,16},
  scaled x ticks=false,
  xticklabel style={/pgf/number format/1000 sep={\mathord{,}}},
  clip=false,
]
\path[fill=psaccent!7] (axis cs:966,4) rectangle (axis cs:3800,13.8);
\draw[psaccent!40,densely dotted,line width=0.6pt] (axis cs:966,13.8) -- (axis cs:3800,13.8);
\draw[pspath,->,>=latex,line width=0.9pt,shorten <=3pt,shorten >=4pt]
  (axis cs:0,5.9) -- (axis cs:0,9.9);
\draw[pspath,->,>=latex,line width=0.9pt,shorten <=3pt,shorten >=5pt]
  (axis cs:0,9.9) -- (axis cs:966,13.8);
\node[font=\footnotesize,anchor=east,align=left,text=psaccent,inner sep=1.2pt,font=\scriptsize,text=psaccent!70] at (axis cs:3630,4.9) {Dominated by AdaRSI};
\draw[psaccent!30,dashed,line width=0.45pt] (axis cs:966,6.9) -- (axis cs:966,13.8);
\node[font=\footnotesize,anchor=west,align=left,text=psmuted!85,inner sep=1.2pt,font=\scriptsize] at (axis cs:1130,11.05) {same procedure\\and policies};
\draw[psvariant!65,line width=0.4pt] (axis cs:1204,7.2) -- (axis cs:1430,8.05);
\draw[psfixed!65,line width=0.4pt] (axis cs:966,6.9) -- (axis cs:1390,6.1);
\addplot[only marks,mark=*,mark size=2.5pt,color=psmuted,mark options={draw=white,line width=0.65pt}] coordinates {(0,4.6)};
\addplot[only marks,mark=o,mark size=2.7pt,color=psbase,line width=0.85pt] coordinates {(0,5.9)};
\addplot[only marks,mark=triangle*,mark size=3.1pt,color=psstage,mark options={draw=white,line width=0.65pt}] coordinates {(0,9.9)};
\addplot[only marks,mark=square*,mark size=2.7pt,color=psvariant,mark options={draw=white,line width=0.65pt}] coordinates {(1204,7.2)};
\addplot[only marks,mark=pentagon*,mark size=3.0pt,color=pswarm,mark options={draw=white,line width=0.65pt}] coordinates {(3411,12.6)};
\addplot[only marks,mark=x,mark size=3.0pt,color=psfixed,line width=0.95pt] coordinates {(966,6.9)};
\addplot[only marks,mark=*,mark size=3.7pt,color=psaccent,mark options={draw=white,line width=0.65pt}] coordinates {(966,13.8)};
\node[font=\footnotesize,anchor=west,align=left,text=psmuted,inner sep=1.2pt,] at (axis cs:145,4.6) {GR00T N1.5};
\node[font=\footnotesize,anchor=west,align=left,text=psbase,inner sep=1.2pt,] at (axis cs:145,5.8) {Agent + VLA};
\node[font=\footnotesize,anchor=north west,align=left,text=psstage,inner sep=1.2pt,] at (axis cs:145,9.6) {Agent +\\harness};
\node[font=\footnotesize,anchor=west,align=left,text=psvariant,inner sep=1.2pt,] at (axis cs:1510,8.15) {Agent + weights};
\node[font=\footnotesize,anchor=west,align=left,text=psfixed,inner sep=1.2pt,] at (axis cs:1460,5.9) {Fixed procedure};
\node[font=\footnotesize,anchor=north east,align=right,text=pswarm,inner sep=1.2pt,] at (axis cs:3650,11.9) {Always train\\{\scriptsize 12.6\% $\cdot$ 3,411 min}};
\node[font=\footnotesize,anchor=west,align=left,text=psaccent,inner sep=1.2pt,] at (axis cs:1180,14.9) {\textbf{\ourmethod}\\{\scriptsize 13.8\% $\cdot$ 966 min}};
\end{axis}
\end{tikzpicture}
\endgroup
\caption{\textbf{Success against training cost on Composite-Unseen.} Arrows trace \ourmethod, which improves the harness first and then fine-tunes on demand. Systems in the shaded region cost at least as much as \ourmethod and succeed less.}
\label{fig:pareto}
\end{figure}
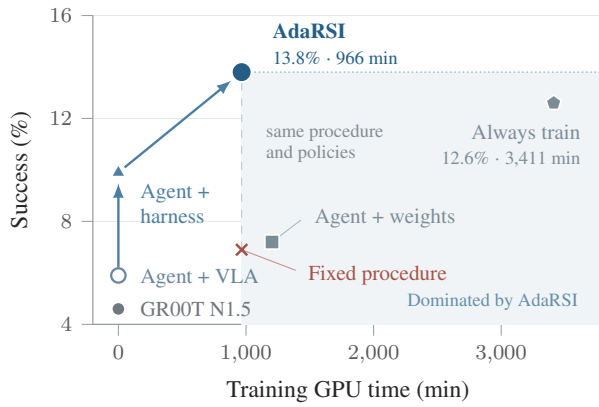

%% file: tables/swap.tex
\begin{table}[t]
\centering
\begingroup
\small
\definecolor{tableaccent}{HTML}{245E86}
\definecolor{tableours}{HTML}{E4EEF8}
\definecolor{tablewarning}{HTML}{A65345}
\definecolor{tablemuted}{HTML}{797979}
\newcommand{\tabaccent}[1]{\textcolor{tableaccent}{#1}}
\newcommand{\tabwarning}[1]{\textcolor{tablewarning}{#1}}
\newcommand{\tabmuted}[1]{\textcolor{tablemuted}{#1}}
\setlength{\tabcolsep}{3pt}
\renewcommand{\arraystretch}{1.18}
\begin{tabularx}{\columnwidth}{>{\raggedright\arraybackslash}m{88pt} CC >{\columncolor{tableours}}C}
\toprule
\textbf{Task} & \makecell{Agent +\\harness} & \makecell{\ourmethod\\w/ $\pi_0$\\[-1pt]{\footnotesize\tabmuted{+ procedure}}} & \makecell{\tabaccent{\ourmethod}\\[-1pt]{\footnotesize\tabmuted{+ weights}}} \\
\midrule
WaffleReheat & 0 & 5\,{\footnotesize \tabaccent{($+5$)}} & \textbf{19}\,{\footnotesize \tabaccent{($+14$)}} \\
RecycleBottlesByType & 0 & 5\,{\footnotesize \tabaccent{($+5$)}} & \textbf{9}\,{\footnotesize \tabaccent{($+4$)}} \\
BreadSelection & 21 & \textbf{27}\,{\footnotesize \tabaccent{($+6$)}} & 26\,{\footnotesize \tabwarning{($-1$)}} \\
WashFruitColander & 24 & 27\,{\footnotesize \tabaccent{($+3$)}} & \textbf{29}\,{\footnotesize \tabaccent{($+2$)}} \\
CuttingToolSelection & \textbf{25} & 20\,{\footnotesize \tabwarning{($-5$)}} & 18\,{\footnotesize \tabwarning{($-2$)}} \\
\midrule
\textbf{Total (of 250)} & \textbf{70} & \textbf{84}\,{\footnotesize \tabaccent{($\mathbf{+14}$)}} & \tabaccent{\textbf{101}}\,{\footnotesize \tabaccent{($\mathbf{+17}$)}} \\
\bottomrule
\end{tabularx}
\endgroup
\caption{\textbf{Weights against procedure.} From left to right, each column adds the revised procedure and then the fine-tuned weights, with the change in parentheses. Entries count successes out of 50 on the five tasks whose success changes in the weight round.}
\label{tab:swap}
\end{table}

%% file: tables/every.tex
\begin{table*}[t]
\centering
\begingroup
\small
\definecolor{tableaccent}{HTML}{245E86}
\definecolor{tablelight}{HTML}{F2F6FA}
\definecolor{tablemuted}{HTML}{444444}
\definecolor{tablewarning}{HTML}{A65345}
\newcommand{\tabaccent}[1]{\textcolor{tableaccent}{#1}}
\setlength{\tabcolsep}{4pt}
\renewcommand{\arraystretch}{1.12}
\begin{tabularx}{\textwidth}{@{}l C >{\columncolor{tablelight}}C C >{\columncolor{tablelight}}C C >{\columncolor{tablelight}}C C >{\columncolor{tablelight}}C C >{\columncolor{tablelight}}C@{}}
\toprule
 & \multicolumn{2}{c}{Successes (of 50)} & \multicolumn{2}{c}{Policies} & \multicolumn{2}{c}{Tokens (M)} & \multicolumn{2}{c}{Demonstrations} & \multicolumn{2}{c}{GPU minutes} \\
\cmidrule(lr){2-3} \cmidrule(lr){4-5} \cmidrule(lr){6-7} \cmidrule(lr){8-9} \cmidrule(l){10-11}
\textbf{Task} & Always & \tabaccent{Ours} & Always & \tabaccent{Ours} & Always & \tabaccent{Ours} & Always & \tabaccent{Ours} & Always & \tabaccent{Ours} \\
\midrule
ArrangeBreadBasket & 0 & \textbf{1} & 5 & \textbf{3} & 184.9 & \textbf{159.8} & 57 & \textbf{17} & 123 & \textbf{57} \\
ArrangeTea & \textbf{1} & 0 & 6 & \textbf{3} & 186.2 & \textbf{161.4} & 72 & \textbf{22} & 164 & \textbf{70} \\
BreadSelection & \textbf{28} & 26 & 9 & \textbf{2} & 163.1 & \textbf{125.2} & 116 & \textbf{17} & 212 & \textbf{56} \\
CategorizeCondiments & \textbf{3} & 1 & 5 & \textbf{2} & 172.6 & \textbf{135.7} & 79 & \textbf{26} & 179 & \textbf{81} \\
CuttingToolSelection & 16 & \textbf{18} & 6 & \textbf{3} & 127.0 & \textbf{108.4} & 89 & \textbf{33} & 163 & \textbf{65} \\
GarnishPancake & \textbf{1} & 0 & 8 & \textbf{3} & 215.7 & \textbf{172.3} & 210 & \textbf{40} & 238 & \textbf{69} \\
GatherTableware & 3 & \textbf{4} & 11 & \textbf{2} & 179.7 & \textbf{118.2} & 218 & \textbf{13} & 345 & \textbf{51} \\
HeatKebabSandwich & 0 & 0 & 6 & \textbf{2} & 207.0 & \textbf{164.1} & 91 & \textbf{30} & 194 & \textbf{63} \\
MakeIceLemonade & 0 & 0 & 6 & \textbf{3} & 180.4 & \textbf{142.4} & 52 & \textbf{18} & 151 & \textbf{56} \\
PanTransfer & 1 & \textbf{2} & 7 & \textbf{2} & 191.1 & \textbf{146.3} & 138 & \textbf{30} & 204 & \textbf{37} \\
PortionHotDogs & \textbf{2} & 0 & 4 & \textbf{1} & 207.1 & \textbf{183.2} & 78 & \textbf{20} & 144 & \textbf{37} \\
RecycleBottlesByType & \textbf{15} & 9 & 7 & \textbf{2} & 181.8 & \textbf{154.3} & 149 & \textbf{24} & 232 & \textbf{57} \\
SeparateFreezerRack & 0 & 0 & 6 & \textbf{3} & 204.7 & \textbf{155.2} & 85 & \textbf{40} & 190 & \textbf{93} \\
WaffleReheat & 14 & \textbf{19} & 10 & \textbf{2} & 274.3 & \textbf{214.1} & 118 & \textbf{24} & 266 & \textbf{53} \\
WashFruitColander & 17 & \textbf{29} & 14 & \textbf{2} & 236.0 & \textbf{185.0} & 163 & \textbf{12} & 383 & \textbf{52} \\
WeighIngredients & 0 & \textbf{1} & 8 & \textbf{3} & 175.5 & \textbf{142.5} & 162 & \textbf{50} & 223 & \textbf{68} \\
\midrule
\addlinespace[2pt]
\textbf{Total} & 101 & \tabaccent{\textbf{110}} & 118 & \tabaccent{\textbf{38}} & 3,086.8 & \tabaccent{\textbf{2,468.0}} & 1,877 & \tabaccent{\textbf{416}} & 3,411 & \tabaccent{\textbf{966}} \\
\addlinespace[2pt]
\textcolor{tablemuted}{Always / Ours} & \multicolumn{2}{c}{\textcolor{tablewarning}{\textbf{0.92$\times$}}} & \multicolumn{2}{c}{\tabaccent{\textbf{3.1$\times$}}} & \multicolumn{2}{c}{\tabaccent{\textbf{1.25$\times$}}} & \multicolumn{2}{c}{\tabaccent{\textbf{4.5$\times$}}} & \multicolumn{2}{c}{\tabaccent{\textbf{3.5$\times$}}} \\
\bottomrule
\end{tabularx}
\endgroup
\caption{\textbf{\ourmethod (Ours) against \emph{Always train} (Always).} Costs count every policy trained, and those of \emph{Always} include the training of \ourmethod. Tokens are input tokens of the improvement phase, and the last row gives Always/Ours ratios.}
\label{tab:every}
\end{table*}

%% file: tables/selection.tex
\begin{table}[t]
\centering
\begingroup
\small
\definecolor{tableaccent}{HTML}{245E86}
\definecolor{tableours}{HTML}{E4EEF8}
\definecolor{tablemuted}{HTML}{797979}
\newcommand{\tabaccent}[1]{\textcolor{tableaccent}{#1}}
\setlength{\tabcolsep}{4pt}
\renewcommand{\arraystretch}{1.13}
\begin{tabularx}{\columnwidth}{l CCC}
\toprule
\textbf{Variant} & \tabaccent{\makecell{Success\\(\%)}} & Demos & \makecell{GPU\\min.} \\
\midrule
Agent + harness & 12.0 & 0 & 0 \\
Lowest first    & 11.8 & 312 & 636 \\
\addlinespace[2pt]
\rowcolor{tableours}
\tabaccent{\textbf{\ourmethod}} & \tabaccent{\textbf{16.6}} & 316 & 796 \\
\bottomrule
\end{tabularx}
\endgroup
\caption{\textbf{Choosing what to train} on the non-pilot Composite-Unseen tasks. \emph{Lowest first} trains the sub-steps on which $\pi_0$ succeeds least, under the same budget.}
\label{tab:selection}
\end{table}

%% file: tables/outcomes.tex
\begin{table*}[t]
\centering
\begingroup
\small
\definecolor{tableaccent}{HTML}{245E86}
\definecolor{tablewarning}{HTML}{A65345}
\definecolor{tablemuted}{HTML}{797979}
\definecolor{tablerule}{HTML}{B5BBC1}
\newcommand{\tabaccent}[1]{\textcolor{tableaccent}{#1}}
\newcommand{\tabwarning}[1]{\textcolor{tablewarning}{#1}}
\newcommand{\tabmuted}[1]{\textcolor{tablemuted}{#1}}
\renewcommand{\arraystretch}{1.13}
\setlength{\tabcolsep}{3pt}
\begin{tabularx}{\textwidth}{@{}>{\raggedright\arraybackslash}m{54pt} >{\raggedright\arraybackslash}m{92pt} >{\centering\arraybackslash}m{26pt} >{\centering\arraybackslash}m{40pt} >{\raggedright\arraybackslash}X@{}}
\toprule
\textbf{Case} & \textbf{Sub-step (task)} & $\pi_0$ & Fine-tuned & \textbf{Lesson} \\
\midrule
\multicolumn{5}{@{}l@{}}{\hbox to \textwidth{\footnotesize\textbf{Training needed}\hspace{6pt}{\color{tablerule}\leaders\hrule height3.3pt depth-3pt\hfill}}} \\
\addlinespace[1pt]
\tabaccent{\textbf{Worked}} & \makecell[l]{Open French-door fridge\\[-1pt]{\footnotesize\color{tablemuted}GarnishPancake}} & 0/6 & \textbf{6/6} & A few demonstrations suffice, but the \textbf{task still fails}. \\
\addlinespace[3pt]
\tabmuted{Partly worked} & \makecell[l]{Pick up the skewer\\[-1pt]{\footnotesize\color{tablemuted}HeatKebabSandwich}} & 0/16 & \textbf{3/16} & The motion is learned in \textbf{only some kitchens}. \\
\addlinespace[3pt]
\tabwarning{\textbf{Did not work}} & \makecell[l]{Close toaster-oven door\\[-1pt]{\footnotesize\color{tablemuted}HeatKebabSandwich}} & \textbf{1/9} & 0/9 & Training \textbf{does not yet produce} the motion, and $\pi_0$ stays the default. \\
\addlinespace[4pt]
\multicolumn{5}{@{}l@{}}{\hbox to \textwidth{\footnotesize\textbf{Training not needed}\hspace{6pt}{\color{tablerule}\leaders\hrule height3.3pt depth-3pt\hfill}}} \\
\addlinespace[1pt]
\tabaccent{\textbf{Helped}} & \makecell[l]{Set up the mug\\[-1pt]{\footnotesize\color{tablemuted}CoffeeSetupMug}} & 13/20 & \textbf{31/40} & Training can \textbf{raise the reliability} of a sub-step $\pi_0$ already performs. \\
\addlinespace[3pt]
\tabmuted{No gain} & \makecell[l]{Turn on tall faucet\\[-1pt]{\footnotesize\color{tablemuted}TurnOnSinkFaucet}} & 11/12 & 11/12 & $\pi_0$ remains the \textbf{better default}. \\
\addlinespace[3pt]
\tabwarning{\textbf{Worse at test}} & \makecell[l]{Open the drawer\\[-1pt]{\footnotesize\color{tablemuted}OpenDrawer}} & 22/27 & \textbf{26/27} & A gain in development \textbf{does not hold at test}, where task success falls. \\
\bottomrule
\end{tabularx}
\endgroup
\caption{\textbf{Outcomes of training a sub-step.} Cases from the pilot tasks, comparing $\pi_0$ and the fine-tuned policy from the same held-out starts. The last three rows come from Atomic-Seen pilots.}
\label{tab:outcomes}
\end{table*}

%% file: figures/case_study_fig.tex
\begin{figure*}[t]
\centering
\includegraphics[width=\textwidth]{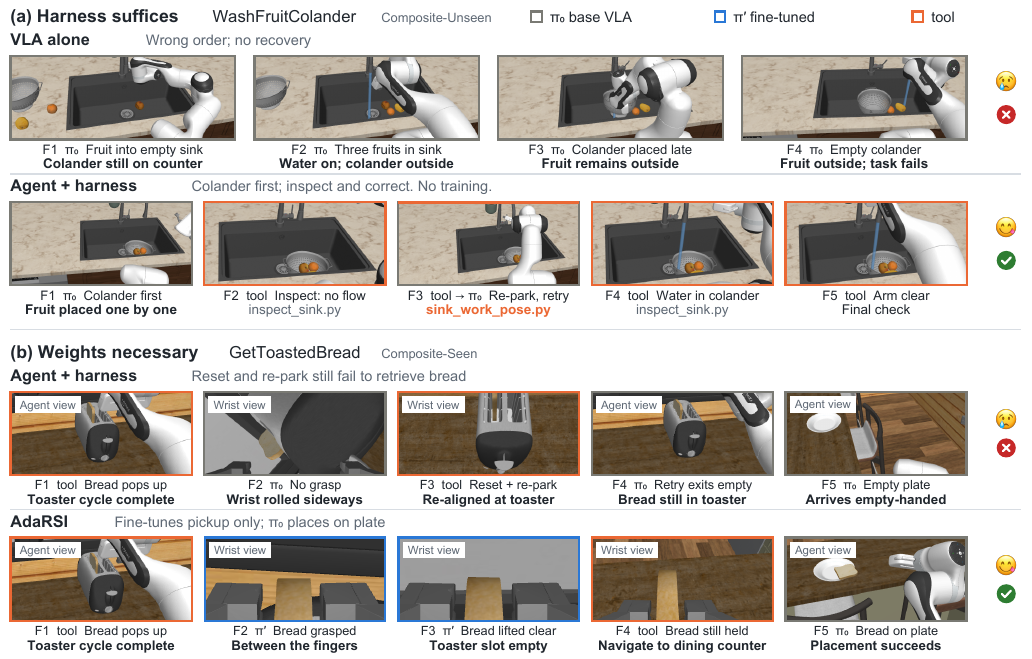}
\caption{\textbf{Case studies.}
(a) WashFruitColander is solved by the harness alone, which fixes the sub-step order and repositions the base.
(b) In GetToastedBread, only the bread pickup is trained.
Rows share the kitchen and initial state.}
\label{fig:case}
\end{figure*}

%% file: sec/6_discussion.tex
\section{Discussion and Limitations}
\label{sec:discussion}

\paragraph{Deciding when to train.}
Our results support a simple principle for embodied self-improvement, to change the harness first and to change the weights only for the sub-steps that stay unreliable once the harness has been refined.
Our instantiation of this principle leaves two decisions to the agent, and it can fail at either.
The agent may find the sub-steps that need training but not obtain a policy that performs them reliably.
This happens in the Composite-Unseen tasks that remain unsolved, which chain several sub-steps beyond the executor and succeed only if every fine-tuned policy in the chain succeeds in the same episode.
The agent may also adopt a change that looks better on a few development scenes but hurts on the test scenes.
In CuttingToolSelection, the weight round starts every episode that needs the peeler with a dedicated grasp step, which looked slightly better in development but halves success in these episodes at test, whether the step runs the fine-tuned policy or $\pi_0$.
Both depend on the capability of the agent and on how training is constrained, including the budget, the number of validation scenes and the way demonstrations are collected.
A more reliable test of necessity and a stronger training loop are natural next steps, and so is restricting the deployed agent to the policies that its procedure selects, which our agent may override after repeated failures.

\paragraph{Limitations.}
All experiments run in simulation, where privileged state makes it easy to generate and label demonstrations.
On a real robot, the same role would have to be played by success detectors, scripted resets or occasional human help, and the cost of a weight change would rise accordingly, which only strengthens the case for training on demand.
We are conducting real-robot experiments and will include them in the next version of this paper.

\section{Conclusion}
We presented \ourmethod, an on-demand recursive self-improvement framework for embodied agents.
It lets a coding agent improve its skills and harness first and fine-tune the executor only on the sub-steps that remain unreliable, using demonstrations it collects itself.
Across all 50 target tasks of RoboCasa365 and without any human demonstration, this improves a pretrained VLA on all three suites, most of all on unseen composite tasks.
The trained weights pay off where the executor lacks a motion, and there they decide whether a task succeeds, whereas fine-tuning a sub-step that the executor already completes often adds cost without adding success and can even make it worse.
Deciding where to train thus lets embodied self-improvement gain from new weights at a fraction of the cost of training everywhere.